# A convex relaxation for weakly supervised classifiers

**Armand Joulin**  ARMAND.JOULIN@ENS.FR
WILLOW-SIERRA project-teams, INRIA - Ecole Normale Supérieure

**Francis Bach**  FRANCIS.BACH@INRIA.FR
SIERRA project-team, INRIA - Ecole Normale Supérieure

## Abstract

This paper introduces a general multi-class approach to weakly supervised classification. Inferring the labels and learning the parameters of the model is usually done jointly through a block-coordinate descent algorithm such as expectation-maximization (EM), which may lead to local minima. To avoid this problem, we propose a cost function based on a convex relaxation of the soft-max loss. We then propose an algorithm specifically designed to efficiently solve the corresponding semidefinite program (SDP). Empirically, our method compares favorably to standard ones on different datasets for multiple instance learning and semi-supervised learning, as well as on clustering tasks.

## 1. Introduction

Discriminative supervised classifiers have proved to be very accurate data-driven tools for learning the relationship between input variables and certain labels. Usually, for these methods to work, the labeling of the training data needs to be complete and precise. However, in many practical situations, this requirement is impossible to meet because of the challenges posed by the acquisition of detailed data annotations. This typically leads to partial or ambiguous labelings.

Different weakly supervised methods have been proposed to tackle this issue. In the semi-supervised framework (Chapelle et al., 2006), only a small number of points are labeled, and the goal is to use the unlabeled points to improve the performance of the classifier. In the multiple-instance learning (MIL) framework introduced by Dietterich & Lathrop (1997), *bags* of instances are labeled together instead of individually, and some instances belonging to the same bag may have different true labels. Finally, in the ambiguous labeling setting (Jin & Ghahramani, 2003; Hullermeier & Beringer, 2006), each point is associated with multiple potential labels.

More generally, in all these frameworks, the points are associated with observable partial labels and the implicit or explicit goal is to *jointly* estimate their true latent labels and learn a classifier based on these labels. This usually leads to a non-convex cost function which is often optimized with a greedy method or a coordinate descent algorithm such as the expectation-maximization (EM) procedure. These methods usually converge to a local minimum, and their initialization remains an open practical problem.

In this paper, we propose a simple and general framework which can be used for any of the aforementioned problems. We explicitly learn the true latent label and the classifier parameters. We also propose a convex relaxation of our cost function and an efficient algorithm to minimize it. More precisely, we use a discriminative classifier with a soft-max loss, and our convex relaxation extends the work of Guo & Schuurmans (2008). Our main contributions are:

- a full convex relaxation of the soft-max loss function with intercept, which can be applied to a large set of multiclass problems with any level of supervision,
- a novel convex cost function for weakly supervised and unsupervised problems and,
- a dedicated and efficient optimization procedure.

We develop our framework for the general weakly supervised case. We propose results on both toy examples as proof of concept of our claims, and on standard MIL and semi-supervised learning (SSL) datasets.

### 1.1. Related work

**Multiple instance learning** (MIL) has received much attention because of its wide range of applications. First used for drug activity prediction, it has also been used in the vision community for different problems such as scene classification (Maron & Ratan, 1998), object detection (Viola et al., 2006), object tracking in video (Babenko et al., 2009), and image database retrieval (Yang, 2000). Many MIL methods have been developed in the past decade. For example, some are based on boosting (Auer & Ortner, 2004), others on nearest neighbors (Wang & Zucker, 2000), on neural networks (Zhang & Zhou, 2006), on decision trees (Blockeel et al., 2005), or the construction of an appropriate kernel (Wang et al., 2008; Gärtner et al., 2002; Kwok & Cheung, 2007). Much of the work in the MIL community has focused on the use of discriminative classifiers, the most popular one being the support vector machine (SVM) (Andrews et al., 2003; Chen & Wang, 2004; Gehler & Chapelle, 2007). In this paper, we concentrate on the logisitic loss which makes little difference with the hinge loss with the additional advantage of being twice differentiable. Note that this loss has already been used in the context of MIL (Xu & Frank, 2004; Ray & Craven, 2005), but with different optimization schemes.

Many **semi-supervised learning** (SSL) methods have also been proposed in the past decade (see, e.g., Chapelle et al., 2006; Zhu, 2006). For example, some are based on maximizing the margin with an SVM framework (Joachims, 1999; Bennett & Demiriz, 1998; Xu & Schuurmans, 2005), and others use the unlabeled data for regularization (Belkin et al., 2004) or co-training of weak classifiers (Blum & Mitchell, 1998).

**Discriminative clustering** provides a principled way to reuse existing supervised learning machinery while *explicitly* estimating the latent labels. For example, following the SVM approach of Xu et al. (2005), algorithms using linear discriminant analysis (De la Torre & Takeo, 2006) or ridge regression (Bach & Harchaoui, 2007) have been proposed. These methods often fail in the multiclass case, whereas we show that the soft-max loss *with intercept* works well in this setting. A common issue for discriminative clustering is that a perfect separation is reached by assigning the same label to all of the points. In most of the previously cited methods, this issue is adressed by adding linear constraints on the size of each cluster. In this paper we use instead a natural cluster-size balancing term corresponding to an entropy penalization (Chapelle et al., 2006; Joulin et al., 2010).

The link between SSL and MIL has been widely studied in the community. For example, in the context of image segmentation with text annotation, Barnard et al. (2003) propose a general weakly supervised model based on a multi-modal extension to a mixture of latent Dirichlet allocation. An important issue with this family of generative models is that learning the parameters is often untractable. Another example is Zhou & Xu (2007) who use the relation between MIL and SSL to develop a method for MIL.

The idea of using a convex cost function in the weakly supervision context has been already studied in different contexts such as, for example, ambiguous labeling (Cour et al., 2009) or discriminative clustering (Xu et al., 2005; Bach & Harchaoui, 2007). In this paper, we are interested in the convex relaxation of a general multiclass loss function, i.e., the soft-max loss. Guo & Schuurmans (2008) propose a related relaxation but do not consider the intercept in the linear classifier. We extend their work to the case of linear classifiers with an intercept and show in the experiment section, why this difference is crucial when it comes to classification. Note that by using kernels, we can use non-linear classifiers as well. Also, our dedicated optimization scheme is more scalable than the one developed in Guo & Schuurmans (2008) and could be applied to their problem as well.

## 2. Proposed model

### 2.1. Notations

We suppose that we observe $I$ bags of instances. For $i$ in $\{1, \ldots, I\}$, $\mathcal{N}_i$ is the set of instances in the $i$-th bag, and $N_i = |\mathcal{N}_i|$ is its cardinality. We denote by $N = \sum_i N_i$ the total number of instances. In each bag $i$, an instance $n$ in $\mathcal{N}_i$ is associated with a feature $x_n \in \mathcal{X}$ and a label $y_n$ in $\mathcal{L}$, in certain feature and label space. In this paper, we suppose that this label is common to all the instances of a same bag and explain only partially the instances contained in the bag. We are thus interested in finding a *latent* label $z_n \in \mathcal{P}$ which would give a better understanding of the data. We denote by $P$ and $L$ the cardinalities of $\mathcal{P}$ and $\mathcal{L}$. We also assume that the latent label $z_n$ of an instance $n$ can only take its values in a subset $\mathcal{P}_{y_n}$ of $\mathcal{P}$ which depends on the label $y_n$ of the bag. The variables $y_n$ and $z_n$ are associated with their canonical vectorial representation, i.e., $z_{np} = 1$ if the instance $n$ has a latent label of $p$ and $0$ otherwise. We denote by $z$ the $N \times P$ matrix with rows $z_n$.

**Instance reweighting.** In many problems, a set of instances can be bigger than the other, this is the case for example in a *one-vs-all* classifer where the number of positive instances is often very small compared to the number of negative examples. A side-contribution of this work is to consider explicitly a reweighting of the data to avoid undesired side effects: Each point is associated with a weight $\pi_n \geq 0$ which denotes its importance compared to others. Some examples are the uniform case, i.e., $\pi_n = \frac{1}{N}$ or when bags have to be reweighted, i.e., $\pi_n = \frac{1}{IN_i}$ for $n$ in the bag $i$. We denote by $\pi$ the vector with entries equal to $\pi_n$. Note that $\pi \geq 0$ and $\sum_n \pi_n = 1$.

This setting is very general, so let us now show how it applies to several concrete settings.

**Semi-supervised learning.** Given a set of true labels $\mathcal{P}$ and $N_l$ points with known label, there are $N_l+1$ bags, i.e., one for each labeled point and one for all the unlabeled instances. The set $\mathcal{L}$ is equal to $\mathcal{P}$ plus a label for the unlabeled bag (i.e., $L = P + 1$). The true label of an instance in a positive bag is fixed whereas in the unlabeled bag it can take any value in $\mathcal{P}$.

**Unsupervised learning.** This is an extreme case of the semi-supervised framework with only the unlabeled bag.

**Multiple instance learning.** There are two possible labels for a bag ($L = 2$), i.e., *positive* ($y_n = 1$) or *negative* ($y_n = 0$). The true label $z_n$ of an instance $n$ in a negative bag is necessarily negative ($z_n = 0$) and in a positive bag it can be either positive or negative ($\mathcal{P}_1 = \{0, 1\}$).

**Ambiguous labelling.** Each bag is associated with a set of possible true labels $\mathcal{P}_l$. The set of partial labels is thus the combination of all possible subsets of $\mathcal{P}$, i.e., each label $l \in \mathcal{L}$ represents a subset of $\mathcal{P}$ ($L = 2^P$).

### 2.2. Problem formulation

The goal of a discriminative weakly supervised classifier is to find the latent labels $z$ that minimize the value of a regularized discriminative loss function. More precisely, given some latent label $z$ and some feature map $\phi : \mathcal{X} \mapsto \mathbb{R}^d$ (note that $\phi$ could be explicitly defined or implicitly given through a positive-definite kernel), we train a multi-class discriminative classifier to find the parameters $w \in \mathbb{R}^{P \times d}$ and $b \in \mathbb{R}^P$ that minimize:

$$L(z, w, b) = \sum_{n=1}^{N} \pi_n \ell(z_n, w^T \phi(x_n) + b),$$

where $\ell : \mathbb{R}^P \times \mathbb{R}^P \mapsto \mathbb{R}$ is a loss function. In this paper, we are interested in the multi-class setting where a natural choice for $\ell$ is the soft-max loss function (Hastie et al., 2001). Note that for a given instance $n$, the set of possible true labels depends on the the label $y$ of its bag, our loss function $\ell(z_n, w^T \phi(x_n) + b)$ then takes the following form:

$$-\sum_{l \in \mathcal{L}} y_{nl} \sum_{p \in \mathcal{P}_l} z_{np} \log\left(\frac{\exp(w_p^T \phi(x_n) + b_p)}{\sum_{k \in \mathcal{P}_l} \exp(w_k^T \phi(x_n) + b_k)}\right),$$

where $w_p^T$ is the $p$–th row of $w^T$ and $b_p$ the $p$–th entry of $b$.

**Cluster-size balancing term.** In many unsupervised or weakly supervised problems, a common issue is that assigning the same label to all the instances leads to perfect separation. In the MIL community, this is equivalent to considering all the bags as negative and a common solution is to add a non-convex constraint which enforces at least one point per positive bag to be positive. Another solution used in the discriminative clustering community is to add constraints on the number of elements per class and per bag (Xu et al., 2005; Bach & Harchaoui, 2007). Despite good results, this solution introduces extra parameters and may be hard to extend to other frameworks such as MIL, where a positive bag may not have any negative instances. Another common technique is to encourage the proportion of points per class and per bag to be close to uniform. An appropriate penalty term for achieving this is the entropy (i.e., $h(v) = -\sum_k v_k \log(v_k)$) of the proportions of points per bag and per latent label, leading to:

$$H(z) = \sum_{i \in I} h\left(\sum_{n \in \mathcal{N}_i} \pi_n z_n\right).$$

Penalizing by this entropy turns out to be equivalent to maximizing the log-likelihood of a graphical model where the features $x_n$ explain the labels $y_n$ through the latent labels $z_n$ (Joulin et al., 2010). An important consequence is that the natural weight of this penalty in the cost function is 1, so we do not add any extra parameters.

To avoid over-fitting, we penalize the norm of $w$, leading to the following cost function:

$$f(z, w, b) = L(z, w, b) - H(z) + \frac{\lambda}{2P} \|w\|_F^2,$$

where $\lambda > 0$ is the regularization parameter and the problem thus takes the following form:

$$\min_{\forall n \leq N,\ z_n \in \mathcal{S}_{P_{y_n}}} \min_{w \in \mathbb{R}^{d \times P},\ b \in \mathbb{R}^P} f(z, w, b), \quad (1)$$

where $\mathcal{S}_P = \{t \in \mathbb{R}^P \mid t \geq 0,\ t^T 1_P = 1\}$ is the *simplex* in $\mathbb{R}^P$. To avoid cumbersome double subscripts, we suppose that any instance $n$ in a bag with a label $y_n$ (which is common to the entire bag), has a latent label $z_n$ in $\mathcal{P}$ instead of $\mathcal{P}_{y_n}$.

In the next section we show how to obtain a convex relaxation of this problem.

## 3. Convex relaxation

An interesting feature of the soft-max cost function is its link to the entropy through the Fenchel conjugate (Boyd & Vandenberghe, 2003), i.e., given a $P$-dimensional vector $t$, the log-partition can be written as $\log\left(\sum_{p=1}^{P} \exp(t_p)\right) = \max_{v \in \mathcal{S}_P} \sum_{p=1}^{P} v_p t_p + h(v)$. Substituting in the loss function, the weakly supervised problem defined in Eq. (1) can be reformulated as:

$$\min_{z \in \mathcal{S}_P^N} \max_{q \in \mathcal{S}_P^N} \sum_{i \in I} \sum_{n \in \mathcal{N}_i} \pi_n h(q_n) - H(z) + g(z, q), \quad (2)$$

A convex relaxation for weakly supervised classifiers

| step | Inner loop update | Inner loop duality gap | Outer loop proximal | Outer loop duality gap |
|---|---|---|---|---|
| complexity | $O(N^2)$ | $O(N^2)$ | $O(N^3)$ | $O(N)$ |

Figure 1. Complexity of the different steps in our algorithm.

where $q$ is an $N \times P$ matrix with $n$-th row $q_n^T$, and $g(z, q)$ is equal to:

$$\min_{\substack{w \in \mathbb{R}^{P \times d} \\ b \in \mathbb{R}^P}} \sum_{i \in I} \sum_{n \in \mathcal{N}_i} \pi_n (q_n - z_n)^T (w^T \phi(x_n) + b) + \frac{\lambda}{2P} \|w\|_F^2.$$

Minimizing this function w.r.t. the intercept $b$ leads to an *intercept constraint* on the dual variables, i.e, $(q - z)^T \pi = 0$. The minimization w.r.t. $w$ leads to a closed-form expression for $g$:

$$g(z, q) = -\frac{P}{2\lambda} \mathrm{tr}\big((q - z)(q - z)^T K\big),$$

where $K$ is the positive definite kernel matrix associated with the reweighted mapping $\phi$, i.e., with entries equal to $K_{nm} = \pi_n \phi(x_n)^T \phi(x_m) \pi_m$. The cost function is not convex in general in $z$ since it is the maximum over a set indexed by $q$ of concave functions in $z$. A common way of dealing with this issue is to relax the problem into a semidefinite program (SDP) in $zz^T$. Unfortunately, our cost function does not directly depend on $zz^T$, but a reparametrization in terms of $q$ inspired by Guo & Schuurmans (2008) allows us to get around this technical difficulty.

**Reparametrization in $q$.** We reparametrize the problem by introducing an $N \times N$ matrix $\Omega$ such that $q = \Omega z$ (Guo & Schuurmans, 2008). The *intercept constraint* and the *normalization constraint* on $q$ (i.e., $q 1_K = 1_N$) become constraints over $\Omega$, i.e., respectively $\Omega^T \pi = \pi$ and $\Omega 1_N = 1_N$. Translating the addition of an intercept to a linear classifier into a simple constraint on the columns of $\Omega$ provides a significant improvement over Guo & Schuurmans (2008), as shown in Section 5.1. This reparametrization has the side-effect of introducing a non-convex term in the cost function since the entropies over $q_n$ in Eq. (2) is replaced by an entropy over the $n$–th row of $\Omega z$ which is not jointly concave/convex in $\Omega$ and $z$.

**Tight upper-bound on the entropy.** We show in the supplementary material that the entropy in $q$ can be bounded by a difference of entropy in $\Omega$ and $z$, up to an additive constant $C_0$:

$$\sum_{i \in \mathcal{I}} \sum_{n \in \mathcal{N}_i} \pi_n h(q_n) \leq -\sum_n \pi_n h(\Omega_n) + H(z) + C_0. \quad (3)$$

This upper-bound is tight in the sense that given a discrete value of $z$ (i.e., before the relaxation), the maximum of the left part among discrete values of $q$ is equal to the maximum of the right part among corresponding discrete values of $\Omega$. Note also that the term in $z$ appearing in Eq. (3) cancels out with the entropy term

in Eq. (2). This relaxation leads to the minimizition of the following function of $z$:

$$\max_{\Omega \in \mathcal{O}} -\frac{P}{2\lambda} \mathrm{tr}\big(zz^T (I - \Omega)^T K (I - \Omega)\big) - \sum_n \pi_n h(\Omega_n),$$

where $\mathcal{O} = \{\Omega \mid \Omega 1_N = 1_N, \Omega^T \pi = \pi, \Omega \geq 0\}$. This problem depends on $z$ solely through the matrix $zz^T$, and can thus be relaxed into an SDP in $zz^T$.

**Reparametrization in $z$.** With the change of variable $Z = zz^T$, we have the maximum of a set of *linear* functions of $Z$, which is convex. However, the set $\mathcal{Z}$ of possible values for $Z$ is non-convex since it is defined by:

$$\begin{cases} \mathrm{diag}(Z) = 1_N, \ Z \geq 0, \ Z \succeq 0, \\ \mathrm{rank}(Z) = k - 1. \end{cases} \quad (4)$$

Let us review these constraints:

- In practice, the piecewise-positivity constraint is not necessary and removing it leads to a matrix $Z$ with entries in $[-1, 1]$ since $Z$ is positive semidefinite with ones on the diagonal.
- The rank constraint is the main source of non-convexity, and will be removed, thus leading to a convex relaxation.
- The rest of the constraints defines the *elliptope*:

$$\mathcal{E}_N = \{Z \in \mathbb{R}^{N \times N} \mid \mathrm{diag}(Z) = 1_N, \ Z \succeq 0\}.$$

Note that an additional linear constraint may be needed depending on the considered weakly supervised problem. We give below some examples:

- In the case of MIL, this constraint takes the form of $Z_- = 1_{N_-} 1_{N_-}^T$, where $N_-$ is the number of negative examples, and $Z_-$ is the restriction of $Z$ to the negative bags.
- "Must-not-link" constraints on the instances can be handled: If two bags $i$ and $j$ have labels $y_i$ and $y_j$ such that the set of possible latent labels are dissimilar (i.e., $\mathcal{P}_{l_i} \cap \mathcal{P}_{l_j} = \emptyset$), we can constrain the submatrix $Z_{ij}$ to be equal to 0. These constraints are of particular interest in the case of SSL, where labeled bags with different labels should not be assigned to the same latent label.

In the rest of this paper, we consider the specific cases of SSL, MIL and discriminative clustering:

- In SSL, we can reduce the dimensionality of $Z$: Since all the values of $z$ with a same known label are equal, it is equivalent to replace them by a

- single element in $Z$. Denoting by $N_u$ is the number of unlabeled points, $P$ the number of labels and $N_R = N_u + P$, this is equivalent to considering a matrix $R^T Z R$ instead of $Z$, where $R$ is a $N \times N_R$ matrix whose restriction to the unlabeled bags is the identity and all other entries are zero except for $R_{n(N_u+l)}$ which is equal to 1 if the instance $n$ has a known label $l$.
- In MIL, the same reduction can be done with $P = 1$ and $N_u$ denoting the total number of positive instances.
- Discriminative clustering is similar to SSL with $P = 0$.

By taking into account all of these modifications and by dropping the rank constraint, we replace the non-convex set $\mathcal{Z}$ by the elliptope $\mathcal{E}_{N_R}$, leading to the minimization of $g(Z)$ over $\mathcal{E}_{N_R}$, where $g(Z)$ is equal to:

$$\max_{\Omega \in \mathcal{O}} -\frac{P}{2\lambda}\text{tr}\big(ZR(I-\Omega)^T K(I-\Omega)R^T\big) - \sum_n \pi_n h(\Omega_n). \quad (5)$$

In the next section we propose an efficient algorithm to solve this convex optimization problem.

## 4. Optimization

Since our optimization involves a maximization in our *inner loop*, it cannot be solved directly by a general-purpose toolbox. We propose an algorithm dedicated to our case. In the rest of this paper we refer to the maximization as the inner loop and the overall minimization of our cost function as the *outer loop*.

### 4.1. Inner loop

Evaluating the cost function defined in Eq. (5) involves the maximization of the sum of the entropy of $\Omega$ and a function $T$ defined as:

$$T(\Omega) = -\frac{1}{2\lambda}\text{tr}\big((I-\Omega)R^T ZR(I-\Omega)^T K\big).$$

We use a proximal method with a reweighted Kullback-Leibler (KL) divergence which naturally enforces the point-wise positivity contraint in $\mathcal{W}$, and leads to an efficient Bregman projection with a KL divergence (an *I-projection* to be more precise) on the rest of the constraints defining $\mathcal{W}$. More precisely, given a point $\Omega^0$, the proximal update is given by maximizing the following function:

$$l_D(\Omega) = \text{tr}\big(\Omega^T \nabla T(\Omega^0)\big) - \sum_n \pi_n h(\Omega_n) - L D_\pi(\Omega \| \Omega^0), \quad (6)$$

where $L$ is the Lipschitz constant of $\nabla T$ and $D_\pi$ is a reweighted KL divergence defined as:

$$D_\pi(\Omega \| \Omega^0) = \sum_i \sum_{n \in \mathcal{N}_i} \pi_n \sum_{m=1}^N \Omega_{nm} \log\left(\frac{\Omega_{nm}}{\Omega^0_{nm}}\right).$$

The *I-projection* can be done efficiently with an *iterative proportional fitting procedure* (IPFP), which is guaranteed to converge to the global minimum with linear convergence rate (Fienberg, 1970).

Note that to obtain a faster convergence of the inner loop, we may take advantage of a low-rank decomposition of $K$ and $R^T Z R$ and we use an accelerated proximal scheme on the logarithm of $\Omega$ (Beck & Teboulle, 2009). To control the distance from the optimum $\Omega^*$, we can use a provably correct duality gap which can be computed efficiently (details are in the supplementary material).

### 4.2. Outer loop

The outer loop minimizes $g(Z)$ as defined in Eq. (5) over the elliptope $\mathcal{E}_{N_R}$. Many approaches have been proposed to solve this type of problems (Goemans & Williamson, 1995; Burer & Monteiro, 2003; Journée et al., 2010) but, to the best of our knowledge, they all assume that the function and its gradient can be computed efficiently and put the emphasis on the projection. This is not the case in our problem, and we thus propose a method adapted to our particular setting.

First, to simplify the projection on the $\mathcal{E}_{N_R}$, we replace our cost function $g(Z)$ by its diagonally rescaled version $g_R(Z) = g(\text{diag}(Z)^{-1/2} Z \text{diag}(Z)^{-1/2})$. Note that even if this function is in general non-convex, it coincides with $g(Z)$ on $\mathcal{E}_{N_R}$, making its restriction to this set convex. This modification allows us to rescale the diagonal of any update $Z$ to a diagonal equal to $1_N$ without modifying the value of our cost function.

Our minimization of $g_R$ over the elliptope is also based on a proximal method with a Bregman divergence to guarantee updates that stay in the feasible set. A natural choice for the Bregman divergence is the KL divergence based on the von Neumann entropy, i.e, the entropy of the eigenvalues of a matrix (see more details in the supplementary material). This divergence guarantees that each update has non-negative eigenvalues. Given a point $Z_0$, its update can then be obtained in closed-form as the diagonally rescaled version of $V\text{Diag}(\exp(\text{diag}(\frac{1}{t}E)))V^T$, where $V$ and $E$ are the eigenvectors and the eigenvalues of $-\nabla g_R(Z_0) + t\log(Z_0)$ and $t$ is a positive step size computed using a line-search with backtracking.

As in the inner loop, we use a computationnally tractable provable duality gap, i.e., $-N_R \lambda_{min}$, where $\lambda_{min}$ is the lowest eigenvalue of $\nabla g_R(Z)$ (see details in the supplementary material).

### 4.3. Rounding

Many rounding schemes can be applied with similar performances. Following Bach & Harchaoui (2007)

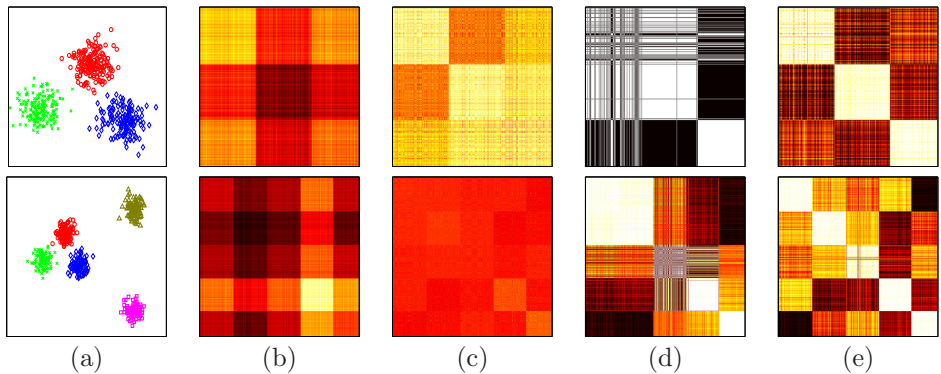

(a) (b) (c) (d) (e)

Figure 2. (a) The clustering problem, (b) the given kernel matrix $K = xx^T$, (c) the matrix $Z$ obtained with (Bach & Harchaoui, 2007), (d) the matrix $Z$ obtained with no intercept and (e) our method (best seen in color).

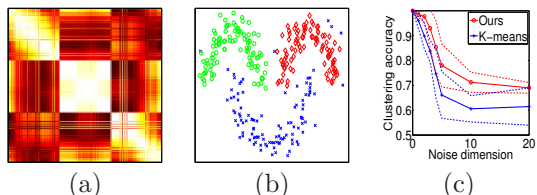

(a) (b) (c)

Figure 3. (a) The matrix obtained with our method and (b) its corresponding clusters. (c) Comparison with k-means on noise robustness ($P = 3$, $N = 300$).

and Joulin et al. (2010), we use k-means clustering on the eigenvectors associated with the $k$ highest eigenvalues (Ng et al., 2001) to obtain a rounded solution $z^*$. This $z^*$ is then used to initialize an EM procedure to solve the problem defined in Eq. (1) and obtain the parameters $(w, b)$ of the classifier, leading to finer details not caught by the convex relaxation.

A specificity of the MIL framework is that strictly no point from a negative bag should be classified as positive, which leads to adding to Eq. (1), the following linear constraints on the parameters of the classifier:

$$\forall i \in I_-,\ n \in \mathcal{N}_i,\ w_0^T \phi(x_n) + b_0 \geq w_1^T \phi(x_n) + b_1. \quad (7)$$

We add these hard constraints in the M-step (optimization over $w$ and $b$) of the EM procedure. The projection over this set of linear constraints is performed efficiently with an homotopy algorithm in the dual (Mairal et al., 2010).

## 5. Results

**Implementation.** Our algorithm is implemented in MATLAB and takes from 1 to 5 minutes for 500 points. Note that we can efficiently compute the solutions for different values of $\lambda$ using warm restarts. Our overall complexity is $O(N^3)$ but we can scale up to several thousands of points. The complexity of the different steps in our algorithm is given in Figure 1. On larger datasets, we can use our relaxation on subsets of instances or on pre-clustering the instances (with k-means) and use it to initialize the EM on the complete dataset.

### 5.1. Discriminative clustering

In this section, we compare our method to two different discriminative clustering methods for the multiclass case: the SDP relaxation of the soft-max problem with no intercept (Guo & Schuurmans, 2008) and the discriminative clustering framework introduced by Bach and Harchaoui (2007). The latter comparison is relevant since they propose a convex cost function based on the square loss with intercept.

We consider in Figure 2, as a proof of concept, two toy examples where the goal is to find 3 and 5 clusters with *linear* kernels and $N = 500$. Even if the clusters are linearly separable, the set of values of $w$ and $b$ which leads to a perfect separation is very small (Figure 2, panel (a)), making the problem challenging. For fair comparison, we test different regularization parameters and show the one leading to the best performances. We show the matrix $Z$ obtained for the three methods as well as the matrix $K = xx^T$ in Figure 2. We see that our method clearly obtains a better estimation of the class assignment compared to the others, showing the importance of both the soft-max loss and the intercept.

In panels (a) and (b) of Figure 3, we also show that our method works with non-linear kernels in a multiclass setting. Finally, in the panel (c) of Figure 3, we show a comparison with k-means as we increase the number of dimensions containing only noise, following the setup of Bach & Harchaoui (2007). Our setting is the 3-cluster problem shown in Figure 2 with an RBF kernel and $N = 300$. We see that our algorithm is more robust than k-means.

### 5.2. Multiple instance learning

In Figure 4, we show some comparisons with other MIL methods on standard datasets (Dietterich & Lathrop, 1997; Andrews et al., 2003) for a variety of tasks:

A convex relaxation for weakly supervised classifiers a drug activity prediction (*musk*), image classification (*fox, tiger* and *elephant*), and text classification (*trec1*).

| Algorithm | Musk1 | Tiger | Elephant | Fox | Trec1 |
|---|---|---|---|---|---|
| Citation k-NN (Wang & Zucker, 2000) | **91.3** | 78.0 | 80.5 | **60.0** | 87.0 |
| EM-DD (Zhang & Goldman, 2001) | 84.8 | 72.1 | 78.3 | 56.1 | 85.8 |
| mi-SVM (Andrews et al., 2003) | 87.4 | 78.9 | 82.0 | **58.2** | 93.6 |
| MI-SVM (Andrews et al., 2003) | 77.9 | **84.0** | 81.4 | **59.4** | 93.9 |
| PPMM Kernel (Wang et al., 2008) | **95.6** | 80.2 | 82.4 | **60.3** | 93.3 |
| Random init / Uniform | 71.1 | 69.0 | 74.5 | **61.0** | 81.3 |
| Tandom init / Weight | 76.6 | 71.0 | 74.5 | **59.0** | 84.4 |
| No intercept / Uniform | $75.0 \pm 19.5$ | $67.8 \pm 10.4$ | $77.3 \pm 9.2$ | $51.3 \pm 6.4$ | $87.5 \pm 5.2$ |
| No intercept / Weight | $77.8 \pm 15.7$ | $71.0 \pm 10.8$ | $78.9 \pm 9.8$ | $52.1 \pm 5.0$ | $87.3 \pm 5.6$ |
| **Ours / Uniform** | $84.4 \pm 14.0$ | $73.0 \pm 8.2$ | $86.7 \pm 3.5$ | $57.5 \pm 5.9$ | $93.0 \pm 4.7$ |
| **Ours / Weight** | $87.7 \pm 13.3$ | $78.0 \pm 5.4$ | $83.9 \pm 4.2$ | $62.5 \pm 6.4$ | $89.0 \pm 6.2$ |

*Figure 4.* Accuracy of our approach and of standard methods for MIL. We evaluate our method with and without the intercept and with two types of weights. In bold, the significantly best performances.

| | Dataset | Linear | Nonlinear | Entropy-Reg. | **Ours (Linear)** | **Ours (Nonlinear)** |
|---|---|---|---|---|---|---|
| | Digit1 | 79.41 | 82.23 | 75.56 | **$84.57 \pm 0.67$** | $75.45 \pm 2.88$ |
| | BCI | 49.96 | **50.85** | **52.29** | **$52.22 \pm 1.13$** | **$50.21 \pm 1.09$** |
| l=10 | g241c | 79.05 | 75.29 | 52.64 | **$87.15 \pm 0.21$** | **$87.29 \pm 0.42$** |
| | g241d | **53.65** | 49.92 | **54.19** | **$54.44 \pm 9.09$** | **$53.15 \pm 10.09$** |
| | USPS | 69.34 | 74.80 | **79.75** | $57.08 \pm 13.34$ | **$79.48 \pm 0.50$** |
| | Digit1 | 81.95 | **93.85** | 92.72 | $91.24 \pm 1.66$ | **$93.31 \pm 0.97$** |
| | BCI | 57.33 | 66.75 | 71.11 | **$78.12 \pm 2.26$** | $64.04 \pm 0.87$ |
| l=100 | g241c | 81.82 | 81.54 | 79.03 | **$86.02 \pm 0.72$** | $85.13 \pm 0.71$ |
| | g241d | **76.24** | **77.58** | 74.64 | **$77.11 \pm 1.65$** | $73.03 \pm 3.02$ |
| | USPS | 78.88 | **90.23** | 87.79 | $71.62 \pm 2.62$ | $73.04 \pm 0.19$ |

*Figure 5.* Comparison in accuracy on SSL databases with methods proposed in (Chapelle et al., 2006). In bold, the significantly best performances.

For comparison, we use the setting described by Andrews et al. (2003), where we create 10 random splits of the data, train on 90% of them and test on the remaining 10%. We test our algorithm with and without the intercept and with uniform or bag-specific (i.e., $\frac{1}{IN_i}$ for instances in the bag $i$) weights, and compare it to some classical MIL algorithms. Note that we have only tried a linear kernel, and we select the regularization parameter using a 2-fold cross-validation for each split. Our algorithm obtains comparable performances with methods dedicated to the MIL problem.

### 5.3. Semi-supervised learning

For the SSL setting, we choose the standard SSL datasets and we compare with methods proposed in Chapelle et al. (2006). The benchmarks (Linear and Nonlinear) are based on a SVM formulation and the benchmark (Entropy-Reg.) uses an entropy regularization. We use our method with either a linear or a RBF kernel. To fix our parameters, we follow the experimental setup of Chapelle et al. (2006). Each set contains 1500 points and either $l = 10$ or $100$ of them are labeled. We show the results in Figure 5. As expected, since the benchmarks and our formulation are very related, the performances are mostly similar when $l = 100$. However, when $l = 10$, our method is more robust and its performances get significantly higher showing that a convex relaxation is less sensible to noise and poorly labeled data.

## 6. Conclusion

In this paper, we propose a convex relaxation of a general cost function for weakly supervised problems. We show the importance of a tight convex relaxation compared to relaxation where either the related linear classifier has been approximated (absence of intercept) or the loss function (square-loss instead of the softmax loss). Our comparison with standard *non-convex* methods for MIL and SSL shows the importance of the initialization for robustness of the approach. We believe that convex relaxation is a powerful tool to obtain good initializations to non-convex problems. The trade-off is that these methods are usually not scalable which suggest to use them on subsets of points or after a quantization step to initialize a more efficient algorithm, such as EM.

**Acknowledgements.** This paper was partially supported by the European Research Council (SIERRA and VIDEOWORLD projects).

# A convex relaxation for weakly supervised classifiers